# Phrase database Approach to structural and semantic disambiguation in English-Korean Machine Translation

Myong-Chol Pak
Foreign Language Faculty
**Kim Il Sung** University
Pyongyang, Democratic People's Republic of Korea

**ABSTRACT**

In machine translation it is common phenomenon that machine-readable dictionaries and standard parsing rules are not enough to ensure accuracy in parsing and translating English phrases into Korean language, which is revealed in misleading translation results due to consequent structural and semantic ambiguities.

This paper aims to suggest a solution to structural and semantic ambiguities due to the idiomaticity and non-grammaticalness of phrases commonly used in English language by applying bilingual phrase database in English-Korean Machine Translation (EKMT). This paper firstly clarifies what the phrase unit in EKMT is based on the definition of the English phrase, secondly clarifies what kind of language unit can be the target of the phrase database for EKMT, thirdly suggests a way to build the phrase database by presenting the format of the phrase database with examples, and finally discusses briefly the method to apply this bilingual phrase database to the EKMT for structural and semantic disambiguation.

## 1. INTRODUCTION

We live in an information society. The IT age offers easy access to variety of information in English. But the language barrier interferes with this access. The attempts to remove this barrier by developing the machine translation systems from English to Korean have been made in the D.P.R.K. since 1990s. However, the translation quality of English-to-Korean machine translation (EKMT) system fails to satisfy the users' needs.

The major reason lies in the several difficulties, which bring about the ambiguities in parsing and selecting the correct Korean language, due to the differences between English and Korean, so overcoming these difficulties is the key point for the success of EKMT system development.

In this paper, I am going to discuss one method to solve this problem by applying English-Korean phrase database approach, that is, how to build the phrase database and how to apply to it to the MT system.

## 2. WHAT IS PHRASE UNIT?

The term "phrase unit" may sound like a new concept in EKMT development.

In general, MT system uses several linguistic databases such as dictionaries, parsing rules,

several kinds of corpus and so on. Here, the dictionary, as the basic linguistic database, refers to the machine-readable dictionary, which is used at several steps in the process of MT, such as morphologic analysis, POS (Part of Speech) tagging, parsing, and synthesizing process. And the parsing rules have been built by the developers as the linguistic database for analyzing structure of the input sentences, which is the main process of MT, and they are of little difference depending on the translation methods of the MT systems.

The general tendency in EKMT system development is that the system, no matter what methods it uses, still fails to overcome inaccuracy in its parsing and translating numerous English phrases only with the machine-readable dictionaries and the built-in parsing rules, and this leads to structural and semantic ambiguities, causing poor translation result.

For example, according to the conventional parsing rules and dictionaries, the English phrase "*call in the loan*" as in "*The bank can call in the loan at any time.*" is interpreted literally as it means "*visit loan*". This doesn't make any sense. The phrasal meaning of "*call in the loan*" is actually "*to officially tell someone to pay back money you lent them*", which illustrates the limitation of the standard parsing rules and machine dictionaries.

Take one more example. The English verb phrase "*take it as read*" in "*It isn't official yet, but you can take it as read that you've got the contract*" must also be parsed and translated as one part of sentence which means "*to feel certain that something is true although no one has told you it is true*". But if this phrase is parsed in separate words as "take", "it", "as", and "read", it causes the POS tagging error, parsing error, and even Korean synthesizing error, and as a result the output semantically contradicts the intention of the writer.

A lot of examples like these can be found in English language.

As one method to solve these problems, we have introduced the idea of bilingual phrase database for structure and word sense disambiguation.

In Longman Dictionary of Contemporary English, the word "PHRASE" is defined as *a group of words that together have a particular meaning, especially when they express the meaning well in a few words*.

And in Oxford Advanced Learner's Dictionary, it is defined as follows;
- a group of words without a finite verb, especially one that forms part of a sentence
- a group of words which have a particular meaning when used together

As we can see above the definitions of the word "phrase" in the dictionaries, the English phrase can be defined as a grammatical unit and a fixed expression. In other words it is a string of words that forms one grammatical unit, usually within a clause or sentence, and a string of words that is used together and has an idiomatic meaning. And many ambiguities are caused by these English phrases, because they are not successfully parsed with the standard grammatical parsing rules.

Based on the definitions of the English phrase given above, the phrase unit that is to be built in EKMT phrase database can be defined as the word combination which functions as one part of sentence with one semantic unit, and its range covers from noun and verb to clause and sentence.

Its meaning is not the semantic sum of the meanings of the individual words which the phrase consists of, and the Korean equivalent for a given phrase unit will also be in the form of phrase unit in the Korean language. In other words this phrase unit refers to the language unit that cannot be parsed or translated with accuracy by the standard grammar rules. Therefore it must be treated as a single unit like an independent word for parsing and translating.

We can classify the phrase unit into various forms such as NOUN phrase unit, VERB phrase unit, ADJECTIVE phrase unit, PREPOSITION phrase unit and so on. Larger units like sentence and clause can be involved into the phrase unit, because structural and semantic ambiguities caused by idiomaticity of some English sentences (Time flies like an arrow, Good bye, How are you?, Happy new year etc.) and clauses (cannot say for certain, as if in a dream, all the indications are that…, etc.) constitute major reason for parsing failure and poor translation result.

### 3. WHAT ARE THE BENEFITS OF THE PHRASE DATABASE?

The major benefit of the phrase database, in a nutshell, is that it greatly contributes to overcoming the deficiency of the existing MT system.

Firstly, the phrase database enlarges the language database, and thus makes the MT translation close to the human translation.

It is self-evident that when the bilingual database is ready, it is easy to translate one language into another. Many English words have multiple meanings, and so do the Korean words. It is very difficult to choose the right Korean equivalent with the exact meaning for a certain English word in machine translation. As a solution to this difficulty we have built several language corpora such as collocation corpus, sense corpus etc. The phrase database, along with these corpora, will ensure much more accuracy in the translation result.

Secondly, the phrase database decreases the dependence on the parsing system based on the POS (Part of Speech) tagging.

Many machine translation programs are rule-based, and it means the accurate POS tagging has a great deal of impact on the process of machine translation, such as parsing, transferring, and synthesizing. The wrong POS tagging will inevitably result in unsatisfactory parsing performance, and even though the POS tagging is successful, the phrase unit will not be parsed precisely due to its characteristics of idiomaticity and non-grammaticalness.

Thirdly, the phrase database ensures flexibility in processing certain phrases.

Before we applied the phrase database, it took long time to parse the English phrase and to choose its Korean equivalents. As we have seen above, the phrase unit includes many idiomatic phrases (*call in the loan, take it as read, cannot say for a certain*) and sentences like proverbs (*Time flies like an arrow*), conversational talks (*How are you, Good bye!*), which are not easy to translate correctly into Korean. In the each process of machine translation such as POS tagging, parsing, and so on, it is usual that the machine translation system is burdened with these phrase units for their non-standard structures and idiomatic Korean equivalents.

And finally, the phrase database makes it easier to enrich the large language database in terms of time and efforts.

In general, it takes a lot of time to build the language database large enough for machine translation system. For example, we have built the English-Korean sentence corpus as large as 150 thousands for 1 year, the English-Korean collocation corpus as large as 4 million words for 2 years, and the Korean sentence corpus as large as 100 million words for 3 years, and we cannot estimate how much we should build them. Even though it took several years and much skillful work to build these corpora, their efficiencies proved to be not very high. Of course, the amount of time it took to build a corpus does not decide its efficiency. By comparison, it took 6 months to build the phrase database as large as 50 thousands, by which the quality of machine translation result turned out to be considerably improved.

The format of the phrase database is not so complicated that any user can manage it, exerting contributive influence over each process of the machine translation system.

## 4. HOW IS THE PHRASE DATABASE BUILT?

The first step to build the phrase database is to set up the format of the database.

In setting up the format for the phrase database, it is important to ensure simplicity for the building convenience and to establish the specified parameters enough for building the phrase database. Here, the simplicity means we don't need any specially trained builders. In other words any user can build this phrase database and thus the building speed and the size of the database can be easily attained. And the specified parameter means that the builder can write easily the English phrases correctly. In other words he can make any English phrase as the component of the phrase database.

Below is the main format of the phrase database.

```
"English Entry"
[part of speech]
# item1 item2 … item n (GRAMMATICAL CATEGORY): Korean Equivalent

e.g. 1.
    "call"
    [verb]
    # call in a loan (VERB) : 상환을 요구하다
```

The ENTRY must have not space, in other words, it must be one single word.

The PART OF SPEECH tags are NOUN, VERB, ADJ, ADV, PREP, and CONJ. This means that the entry words must be noun, verb, adjective, adverb, preposition and conjunction.

The tags for GRAMMATICAL CATEGORY are PP, VERB, ADV, ADJ, NOUN, and SENT.

Each phrase begins with the symbol "#", and the colon ":" is used as the separation tag between English phrase and its Korean equivalent. The grammatical category should be before the colon ":" in capital letter.

One of the main characteristics of the phrase unit is that it has its own parameters.

Below is the complete table of the parameters set for the phrase database, and they facilitate representing diverse linguistic phenomena.

| Parameter | Meaning |
|---|---|
| NOUN | noun |
| PRON | pronoun |
| PRPN | proper noun |
| NUM | number |
| ONE_S | possessive case |
| ONESELF | reflexive pronoun |
| A, B, C | noun phrase |
| ADJ | adjective |
| ADV | adverb |
| VP | verb phrase |
| PP | preposition phrase |
| SENT | sentence |
| THAT_CLAUSE | that-clause |
| WHAT_CLAUSE | what-clause |
| WHETHER_CLAUSE | whether-clause |
| IF_CLAUSE | if-clause |
| HOW_CLAUSE | how-clause |
| WHERE_CLAUSE | where-clause |
| WH_CLAUSE | wh-clause |
| TO_INF | to-infinitive |
| BARE_INF | bare infinitive |
| PASTP | past participle |
| PRESP | present participle |

Table 1. The parameters for the phrase database

If one phrase unit has more than two identical parameters, the cardinal number is added to each parameter. If one phrase unit has more than two noun phrases, the alphabet letters (A, B, C etc.) will distinguish them.

```
e.g. 2
    "take"
    [verb]
    # it take A for B TO_INF    (SENT) : B 가 TO_INF 하는데 A 가 걸리다
```

One of the purposes for which we use the parameter in building the phrase database is for extending the coverage of the phrase for the input sentences by matching them with any sentences and patterns that have the same or similar phrase unit.

We use different symbols for the representation of the phrase unit, such as <^> (possible for morphological change), <@> (main word), <|> (impossible for modifying in front of the phrase), <$> (impossible for modifying after the phrase), <*> (details for variables), </> (alternative word) and so on.

```
e.g. 3
    "take"
    [verb]
    # have ONE_S [picture/photo] @taken (VERB) : ONE 의 사진을 찍다
    # take a [step/walk/stroll] (VERB) : 산보하다
    # ^take A NUM1 minute TO_INF (VERB) : A 가 TO_INF 하는데 NUM1 분 걸리다
    # take A apart $ (VERB) : A 를 분해하다
    # take it for granted THAT_CLAUSE (VERB): 응당 THAT_CLAUSE 할것으로 생각하다
```

The machine-readable dictionary of the machine translation program "Ryongnamsan" (2.0) is the language database that contains the Korean language pattern for the individual English word.

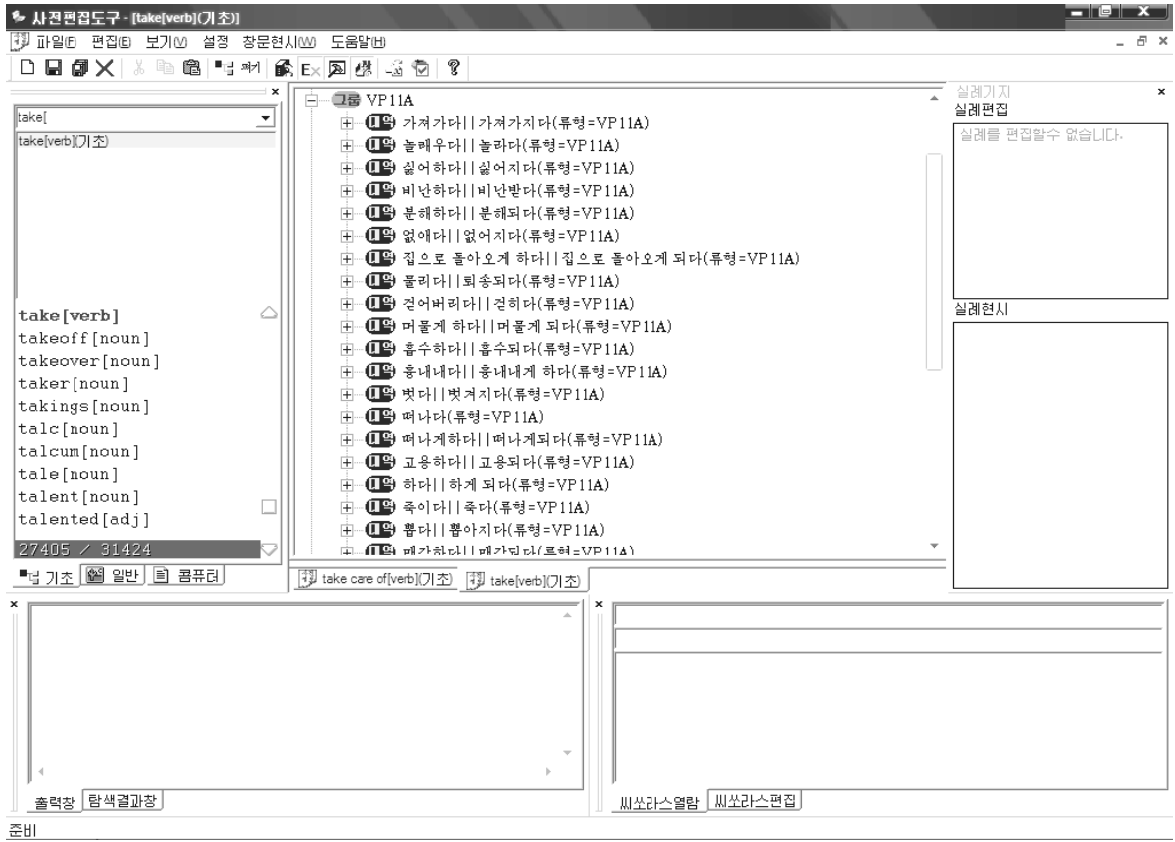

Fig 1. The editor of the dictionary of machine translation program "Ryongnamsan"

In this dictionary each entry is given the translation patterns with the information of part of speech, Korean language, and some detailed information for parsing and synthesizing. And when a sentence is input into the MT system, these information are used in every processes of translation. But these translation patterns are fixed patterns, and these patterns cannot be applied to the phrase unit.

Each language database has its own advantage and disadvantage.

The machine-readable dictionary uses the structurally fixed translation patterns that enable the system to process diverse linguistic phenomena of English sentences inclusively. But it is weak at the description of the idiomatic and non-grammatical expressions that are commonly used in English sentences. By comparison, the phrase database has no restrictive rules on description of all kinds of English expressions including idiomatic and non-grammatical ones. But it can be applied only when the matching condition is provided. In other words it has a lack of ability to process the transformed linguistic phenomena of English language, particularly when the verbs are in passive forms.

## 5. WHAT DOES THE PHRASE DATABASE INCLUDE?

The principles that we adhere to building the phrase database are as follows:

Firstly, we have not included the English phrases that can be parsed by the standard grammatical parsing rules. It is because the idea of using phrase database in machine translation programme is not for the standard grammatical expressions but for the idiomatic and non-grammatical ones (e.g. *take it as read*) that cannot be parsed by the grammar rules.

Secondly, we have included the English phrases that have the Korean equivalents different from the parsing result.

In machine translation programme, parsing is very important step, but it is not for its own sake. Its aim is to output the semantically correct Korean sentences. So, if the expression is correctly parsed by the standard grammatical rules and the semantic sum of individual words is equivalent enough to its Korean language unit, it is not necessary to include that kind of phrases in the phrase database. But the English word like "number" which represents amount or quantity is used as a phrase for "a number of" and its Korean language is semantically different from its sum of "a", "number", and "of". So, the phrases like "*a number of*" should be the target component of the phrase database.

Thirdly, we have included the phrases that have the standard translation patterns.

As mentioned in the previous section, the machine-readable dictionary of the machine translation program "Ryongnamsan" (2.0) is the language database that contains the fixed translation patterns of the words and their Korean language according to the parts of speech. Particularly, as for the verbs, they are divided into 41 patterns, of which the verbs from VP1 to VP6 are intransitive verbs, and the verbs from VP7 to VP41 are transitive verbs, and the parsing and synthesizing are carried out according to these patterns. And the words like nouns and adjectives have their own patterns.

Fourthly, we have included the English noun combinations in which each component functions as the multi-parts of speech, such as the word combination "*command control*".

It is important to parse the base noun phrase (BNP) correctly in machine translation. Especially, the word combination as noun often causes ambiguity in POS tagging because the noun word in English has the multi parts of speech, such as verb and noun. For example, the word combination "command control" may be one entry as a noun in the machine readable dictionary. But in the sentence "*The trap command controls signals coming into this program while it is running*" the word "control" must be POS tagged as a verb. If not, the parsing will not turn out to be successful. It is common particularly in the word combinations which consist of two words such as *command display, process check, call graph, system need, file request, session end, system support, analysis list, execute permission, terminate module, include file, enable condition,* etc. Thus, such word combinations as noun phrases must be included in the phrase database, and at the morphological analyzing step, they are separated into individual words for POS tagging, after which they are processed based on the phrase database.

And finally, we have included the conversational sentences and idiomatic phrases in the phrase database.

Many conversational sentences (*Good morning*, *How are you*?), proverbs (*Time flies like an arrow.*) and idiomatic phrases (*call in the loan, take it as read, cannot say for a certain*) cannot be parsed by the standard parsing rules, and so are the Korean language for them. But they can be processed on the phrase database.

## 6. WHEN CAN WE APPLY THE PHRASE DATABASE TO MT SYSTEM?

After building the phrase database according to the above principles, the next step is how to apply the phrase database to machine translation system.

First of all, it is important to set the applying points where the phrase database should be used. At early stage, this database was used only at the synthesizing step for the disambiguation in the selected Korean language. But its efficiency was not as desired, and the language resources were not fully used. In other words, this was a waste of the language resources. More effective method should be developed.

The phrase units in this database, as above described, consists of three parts, such as English word string, grammatical information and Korean word string, and these formalization will enable the phrase units to be convertible into the parsing rules.

So we converted the phrase units into the parsing rules and applied them at the parsing step as well. Though these parsing rules from phrase database can be matched only when the whole phrase unit is matched with the phrase of the input sentence, it processes the idiomatic and non-grammatical unit as one linguistic unit so that it can reduce the structural ambiguity.

We applied the phrase database at two steps - first step for one-to-one match and second step for extending match.

- First Step (One-to-One match)
Let's compare the following two tree structures of the example sentence.

e.g.: *You can take it as read that you've got the contract.*

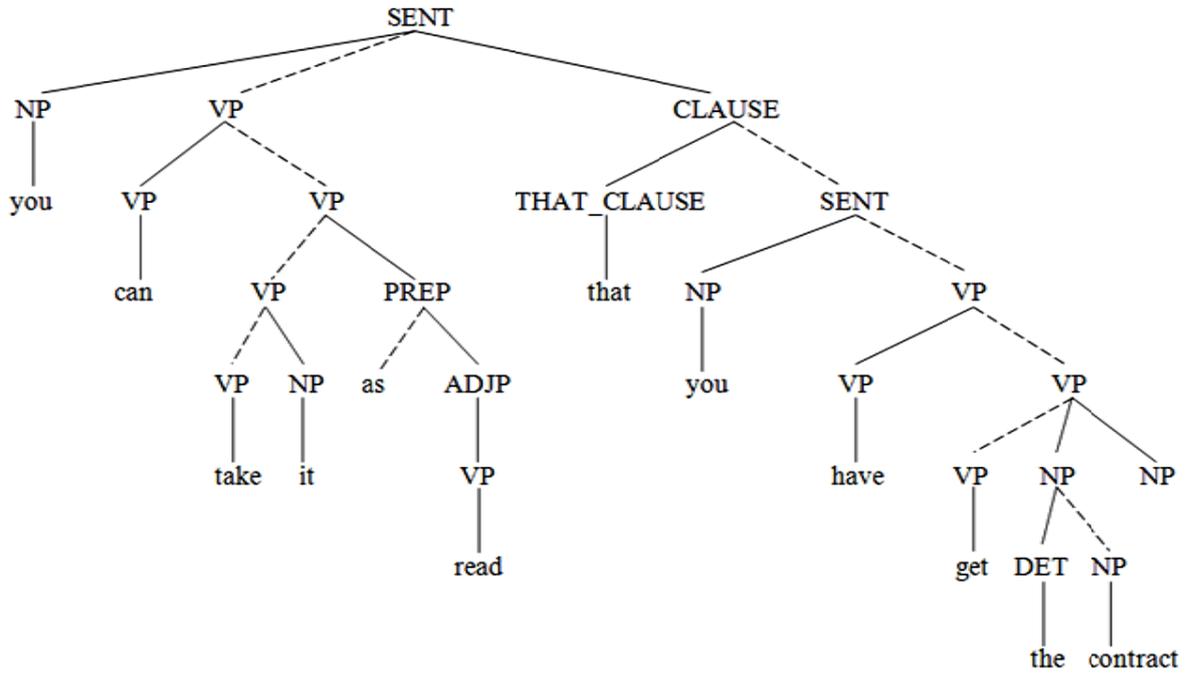

Fig.2. Sentence tree before applying the phrase database

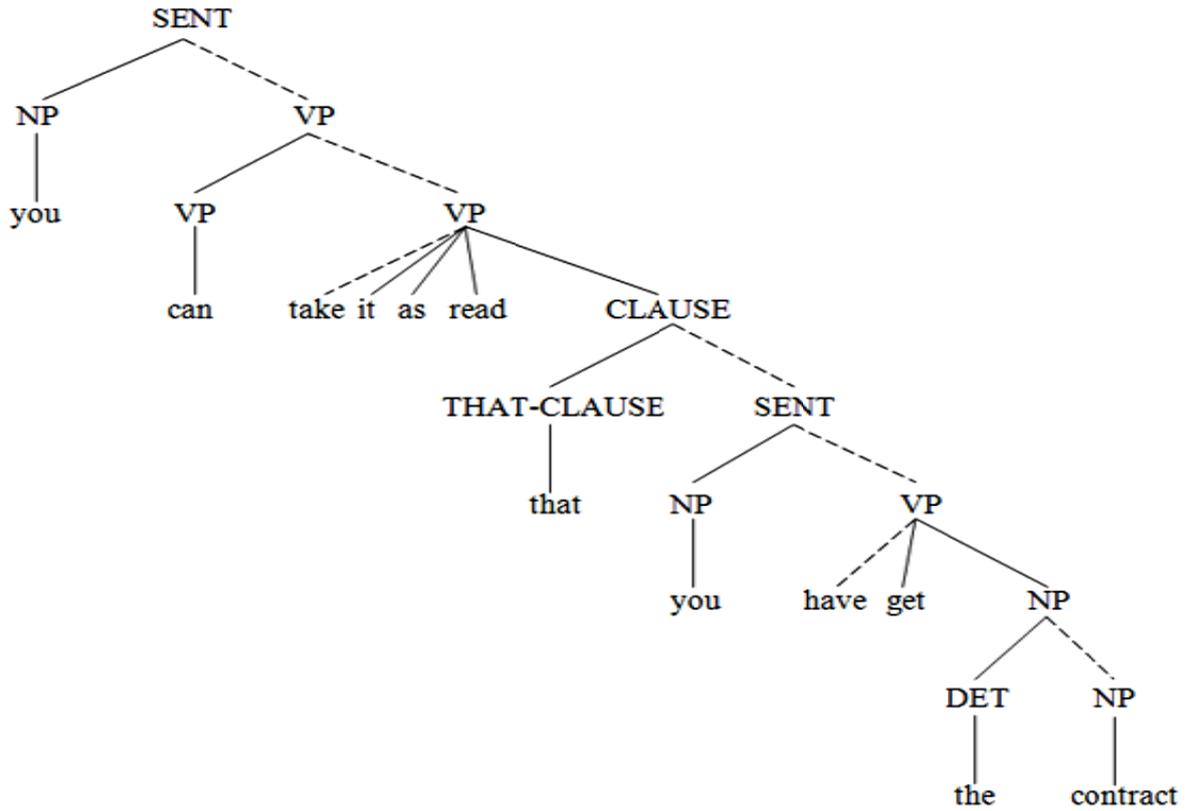

Fig.3. Sentence tree after applying the phrase database

```
PHRASE: "take"
    [verb]
        # take it as read THAT_CLAUSE (VERB) : THAT_CLAUSE 하다고 생각하다
```

As you can see in fig. 3, if the phrase unit is matched with the phrase in the input sentence, we find the matching possible, and the part of speech of the matched phrase in the sentence can be replaced with the grammatical information of the phrase unit in the database. At this step, the structural ambiguity of the input sentence due to the idiomatic and non-grammatical phrase will be removed, and at the synthesizing step the Korean language for the matched phrase of the database is selected as the Korean for that phrase unit. This is called as the phrase unit matching process.

- Second Step (Extending match)

The second step is set for extending the match ability of the phrase units in the database. Sometimes the phrases in input sentences can be extended with some modifiers such as adjective for noun and adverb for verb. In this case it is impossible to build all the extended forms of the phrase units in the database.

For example, in the example sentence "*The bank can call in the loan at any time.*", the matching unit is "*call in the loan*", and the item of this phrase "*loan*" can be extended by adding some adjective such as "*personal*", "*business*", "*interest-free*" and so on. Here, we have to consider the Korean language for the modifiers because in the phrase database, the Korean language for the items of the phrase is not matched to each English word. In order to apply the phrase database at the second step it is essential to, first of all, process the phrase units with modifiers in consideration of their collocation and to select the Korean equivalent to them and identify their location in the Korean language unit. And then it is required to ensure matching condition for the phrase units without modifiers.

## 7. CONCLUSION

By applying the phrase database to the MT system, we have formalized diverse linguistic phenomena of Korean and English languages to provide access to it and made contribution to increasing the capacity of the MT system by fully reflecting the rich translation knowledge of human beings in an easier way and by removing ambiguities in parsing and synthesizing process.

But this methodology still has more to cope with including the problems such as the conflict between phrase units, automatically match between the items of English phrases and Korean equivalents, and so on.

English-to-Korean machine translation program "Ryongnamsan" is still under development, storing the massive phrases in order to improve the translation quality.